# OLiVia-Nav: An Online Lifelong Vision Language Approach for Mobile Robot Social Navigation

Siddarth Narasimhan, *IEEE Student Member*, Aaron Hao Tan, *IEEE Student Member*, Daniel Choi, Goldie Nejat, *IEEE Member*

*Abstract*— Service robots in human-centered environments such as hospitals, office buildings, and long-term care homes need to navigate while adhering to social norms to ensure the safety and comfortability of the people they are sharing the space with. Furthermore, they need to adapt to new social scenarios that can arise during robot navigation. In this paper, we present a novel Online Lifelong Vision Language architecture, OLiVia-Nav, which uniquely integrates vision-language models (VLMs) with an online lifelong learning framework for robot social navigation. We introduce a unique distillation approach, Social Context Contrastive Language Image Pre-training (*SC-CLIP*), to transfer the social reasoning capabilities of large VLMs to a lightweight VLM, in order for OLiVia-Nav to directly encode social and environment context during robot navigation. These encoded embeddings are used to generate and select robot social compliant trajectories. The lifelong learning capabilities of *SC-CLIP* enable OLiVia-Nav to update the robot trajectory planning overtime as new social scenarios are encountered. We conducted extensive real-world experiments in diverse social navigation scenarios. The results showed that OLiVia-Nav outperformed existing state-of-the-art DRL and VLM methods in terms of mean squared error, Hausdorff loss, and personal space violation duration. Ablation studies also verified the design choices for OLiVia-Nav.

## I. INTRODUCTION

Robot social navigation refers to the ability of an autonomous robot to move towards a goal within a human-centered environment while adhering to socially acceptable norms and behaviors [1]. Mobile service robots have been deployed in human spaces for tasks such as delivery of medication in hospitals [2], floor cleaning in office buildings [3], and patrolling for activity monitoring in long-term care facilities [4]. To promote safety and comfortability with humans while conducting such tasks, robots should navigate using social-awareness [5]. This includes respecting personal space [6], interpreting human movement intentions [7], and providing right of way to vulnerable people [8]. However, performing these social-aware actions can be challenging as robots must react to human behavior and contexts in real time [9], while dealing with varying social conditions [10].

Existing robot social navigation approaches have mainly used either human-model-based (HMB) [11]-[18] or human-model-free (HMF) [19]-[24], methods. In HMB methods, human trajectories are explicitly predicted and then incorporated into a navigation policy primarily using deep reinforcement learning (DRL) [11]-[18]. HMF methods implicitly account for human trajectories by: 1) learning social navigation policies using imitation learning (IL) [19]-[22] or 2) leveraging social reasoning capabilities of large foundation models such as large language models (LLMs) [24] or VLMs [23]. However, HMB and HMF methods do not account for social context, such as social scenarios that entail passing conversational groups or navigating against traffic, or environment context, such as open spaces versus narrow hallways. These contexts are important for robot path planning [23]. Furthermore, they are unable to adapt to new social scenarios (unexpected human behaviors, changes in the environment), resulting in degraded performance in real-world deployment [9].

In this paper, we present a novel **O**nline **Li**felong **Vi**sion L**a**nguage architecture, OLiVia-Nav, for mobile robot social navigation, which considers both social and environment context during robot trajectory planning and adapts to new social scenarios. OLiVia-Nav is the first architecture to leverage both the social reasoning capabilities of large VLMs, and the smaller size and faster response time of lightweight VLMs to generate social context embeddings of a robot's surroundings. Our main contributions are: 1) the development of a novel distillation process, *Social Context Contrastive Language Image Pre-training (SC-CLIP),* that transfers the social reasoning capabilities of large VLMs into two lightweight encoders. These encoders extract social context embeddings from both visual and token semantic features from image and text captions, respectively, for predicting robot trajectories, and 2) the development of a trajectory planning network that uniquely utilizes multi-head attention to account for these social context embeddings during the generation of socially compliant robot trajectories. The encoders support online lifelong learning to adapt to new unseen social navigation scenarios during robot navigation.

## II. RELATED WORKS

Existing robot social navigation approaches can be categorized into: 1) human-model-based (HMB) methods [11]-[18], 2) human-model-free (HMF) methods [19]-[24], and 3) lifelong learning methods [25], [26].

### A. Human-Model-Based Methods (HMB)

In general, HMB methods explicitly predict the trajectories of people in the robot's surrounding and use them to inform robot social navigation.

Human trajectories have been predicted using constant velocity (CV) models [16], transformer models [11], [13], [18], or human tracking (HT) models [12], [14], [15] using RGB images and LiDAR point clouds. In CV models, human

This work was supported in part by the Natural Sciences and Engineering Research Council of Canada (NSERC), and in part by the Canada Research Chairs program (CRC). The authors are with the Autonomous Systems and Biomechatronics Laboratory (ASBLab), Department of Mechanical and Industrial Engineering, University of Toronto, Toronto, ON M5S 3G8, Canada (e-mail: {s.narasimhan, jeongwoong.choi}@mail.utoronto.ca, {aaronhao.tan, goldie.nejat}@utoronto.ca; Corresponding author: *Siddarth Narasimhan* trajectories are predicted by assuming a constant speed and direction [16]. In transformer models, a spatial-temporal graph transformer is used to encode distance relationships between people overtime to predict their positions [11], [13], [18]. Lastly, HT models utilize tracker methods such as YOLO [27] to detect and track human positions using either bounding boxes [12], [15] or LiDAR point clouds [14]. These models then predict trajectories by estimating velocity and direction from the tracked human positions.

The predicted trajectories are then used to generate social navigation policies by using DRL methods [11], [13], [14], [16]-[18] or heuristic control methods [12], [15]. Namely, DRL techniques include actor-critic [11], [14], proximal policy optimization (PPO) [13], [16], [17], and double deep Q network [18], which optimize reward functions to maximize personal space and minimize collisions. Heuristic control policy methods consist of predefined rules used to generate policies to avoid human trajectories [12], [15].

The HMB methods are trained using 2D simulated environments, where humans are represented as point masses [11], [12], [14] [16], [18], or 3D simulated environments with procedurally generated human trajectories [13], [17].

### B. Human-Model-Free (HMF) Methods

HMF methods consist of: 1) imitation learning (IL) methods [19]-[22] which learn social navigation policies from expert knowledge in datasets, or 2) large foundation model methods (e.g., LLMs/VLMs) [24], [23]), that generate robot actions based on their social reasoning capabilities.

IL methods use behavior cloning [19], transformer architectures [20], generative adversarial IL [21], or inverse reinforcement learning [22] to predict robot trajectories. They have leveraged social navigation datasets including SCAND [28], MuSoHu [29] and THOR-Magni [30], which contain expert demonstrations of robot trajectories, RGB images, and LiDAR point clouds from real-world environments.

LLMs and VLMs exhibit social reasoning capabilities as they are pre-trained on internet-scale data [31]. These methods generate social navigation policies in two stages. Firstly, LLMs and VLMs are prompted to generate social and environment context captions using audio [24] or image [23] inputs. Secondly, a navigation planner (DRL [24] or the dynamic window approach (DWA) [23]) use these context captions to generate socially compliant navigation actions.

### C. Lifelong Learning Methods

Lifelong learning methods incrementally update navigation model parameters to adapt to new social scenarios overtime, and have only been applied to HMB methods [25], [26]. For example, in [25], human trajectories were first tracked using a Kalman filter (KF) with visual and LiDAR data to estimate their positions and velocities. Lifelong learning was achieved by updating probability distribution maps with this tracking data, predicting human positions relative to the robot at discrete future time steps. In [26], a DRL architecture using PPO was trained with the THOR-Magni dataset [30] to predict human trajectories using gated recurrent units (GRUs). The predicted trajectories were used to quantify deviations from socially acceptable behaviors through a social cost function, which was incorporated into the DRL reward function. The lifelong learning process continuously updated the weights for the navigation policy by retraining the model with new human interaction data collected during navigation.

### D. Summary of Limitations

Existing HMB methods do not account for social and environment context found in real-world scenarios [11]-[18]. Furthermore, lifelong learning methods that use HMB methods rely on predefined human motion models within a KF. These motion models do not incorporate visual cues from the environment or the behaviors of nearby people [25]. They also cannot be directly applied in real-world environments due to their training on simulation-based dataset, creating a sim-to-real gap [26].

HMF methods that use IL [19]-[22] can only handle social scenarios and the environment directly observed in their training datasets. This is a limitation as existing navigation datasets cannot comprehensively encompass every possible social scenario that may occur [28]. LLM and VLM methods [23], [24] are limited by their slow response times resulting in poor social navigation performance, as they cannot adapt to fast human movements leading to collisions [32].

To address the limitations, we propose OLiVia-Nav to: 1) account for social and environment context using *SC-CLIP* to generate social context embeddings. These embeddings encode the robot's surrounding environment, human behavior, and high-level navigations for socially compliant robot trajectory planning and selection; and 2) adapt to new social scenarios by incorporating online lifelong learning through *SC-CLIP*'s lightweight encoders. These encoders are updated during navigation to continuously account for new unseen social scenarios.

## III. ROBOT SOCIAL NAVIGATION PROBLEM

Robot social navigation addresses the problem of a mobile robot that needs to navigate from its initial pose $(x_0, y_0, \phi_0)$ to a goal pose $(x_G, y_G, \phi_G)$ in an unknown environment consisting of dynamic people. The robot uses RGB images from its onboard camera, $I_{RGB}$, to detect visual features such as people and objects, and 3D LiDAR point clouds, $L_{(x,y,z)}$, to provide the 3D structural layout of the robot's surrounding. The task is to predict $K$ socially compliant future trajectories, $\tau^P$, from $(x_0, y_0, \phi_0)$ to $(x_G, y_G, \phi_G)$ given an expert demonstration trajectory $\tau^E = \{(x_j, y_j, \phi_j)\}_{j=0}^G$:

$$\tau^P = \{\tau_k^P : \tau_k^P = \{(x_i^k, y_i^k, \phi_i^k)\}_{i=0}^G, k \in [1, K]\}. \quad (1)$$

The trajectories, $\tau^P$, are generated by a deep neural network, $f_\theta(I, L)$, with learnable parameters, $\theta$. The overall objective is to learn $\theta$ in order to minimize the winner-takes-all (WTA) loss [33] between the predicted and expert trajectories:

$$\theta^* = \underset{\theta}{\arg\min} \, \text{WTA}(f_\theta(I, L), \tau^E). \quad (2)$$

## IV. OLIVIA-NAV ARCHITECTURE

The OLiVia-Nav architecture consists of four main modules, Fig. 1: 1) *Social Context Module* (*SCM*), 2)

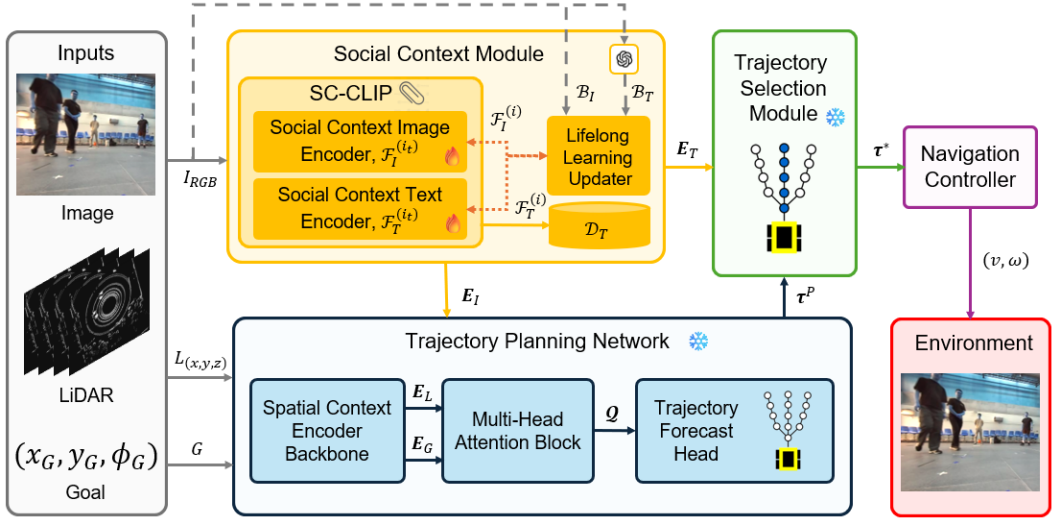

**Fig. 1.** OLiViLa-Nav consists of four modules, 1) *Social Context Module (SCM)* extracts social context embeddings for trajectory planning and selection, and to update the social context image and text encoders for lifelong learning, 2) *Trajectory Planning Network (TPN)* generates socially compliant navigation trajectories using LiDAR data, goal and the social context image embedding, 3) *Trajectory Selection Module (TSM)* selects the trajectory that follows the high-level navigation action encoded in the social context text embedding, and 4) *Navigation Controller (NC)* uses a Proportional Derivative Integral (PID) controller to follow the selected trajectory. 🔥 and ❄ denote modules that are updated and frozen during navigation. 🌀 denotes a VLM.

*Trajectory Planning Network* (*TPN*), 3) *Trajectory Selection Module* (*TSM*), and 4) *Navigation Controller* (*NC*). The *SCM* generates social context embeddings from the robot's surroundings, which are used by the *TPN* and *TSM* to predict and select a socially compliant trajectory for execution by the *NC*. Each module is discussed in detail below.

*A. Social Context Module (SCM)*

The proposed *SCM* consists of two submodules: 1) *Social Context Contrastive Language Image Pretraining (SC-CLIP)*, and 2) a *Lifelong Learning Updater (LLU)*, Fig. 1.

*1) Social Context Contrastive Language Image Pretraining (SC-CLIP)*

We introduce *SC-CLIP*, a distillation approach to transfer the social reasoning of a large VLM to two lightweight encoders: a social context image encoder (*SCIE*), $\mathcal{F}_I$, and a social context text encoder (*SCTE*), $\mathcal{F}_T$. The novelty of *SC-CLIP* is its ability to retain the social understanding of a large VLM, enabling OLiVia-Nav to generalize to diverse social scenarios, without the slow response speed of the large VLM.

The distillation approach consists of two stages. First, a large VLM is used to generate a long and short text caption, $(T_l, T_s)$, to describe the social and environment context within $I_{RGB}$, Fig. 2. This context includes descriptions of: 1) the social scenario, 2) objects and people in the environment, and 3) the high-level navigation action that the robot should follow to remain socially compliant. Note that both $T_l$ and $T_s$ are required in the training procedure of *SC-CLIP* to enable long text caption understanding, as detailed in [34]. Then, in the second stage, *SC-CLIP* trains the $\mathcal{F}_I$ and $\mathcal{F}_T$ in parallel to align the image embedding, $E_I$, from $I_{RGB}$, with the corresponding text embedding, $E_T$, from $(T_l, T_s)$, in their respective embedding spaces. Herein, $\mathcal{F}_I$ utilizes a ViT-L/14 transformer backbone [35] to extract visual semantic features from $I_{RGB}$ in order to generate $E_I$. $\mathcal{F}_T$ utilizes a self-attention transformer backbone [36] to extract token semantic features from $(T_l, T_s)$ to generate $E_T$. *SC-CLIP* is trained using the following loss function [34]:

$$\mathcal{L}_{SC-CLIP} = \text{CE}(\mathcal{F}_I(I_{RGB}) \cdot \mathcal{F}_T(T_l)^T, labels) \\ + \text{CE}(\text{PCE}(\mathcal{F}_I(I_{RGB})) \cdot \mathcal{F}_T(T_s)^T, labels), \quad (3)$$

where PCE is the Principal Component Extraction function, which extracts high-level context features from $\mathcal{F}_I(I_{RGB})$ [37], CE is the cross-entropy loss, and *labels* are the ground truth indices that align $E_I$ and $E_T$.

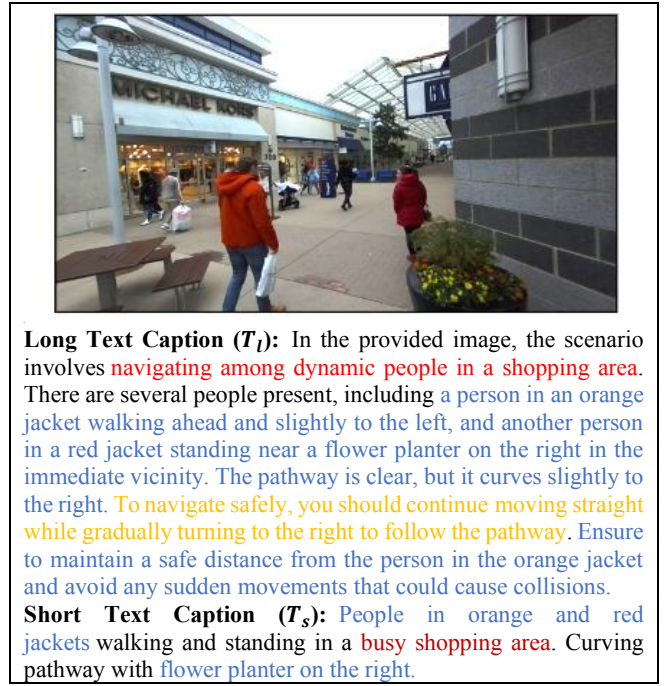

**Long Text Caption ($T_l$):** In the provided image, the scenario involves navigating among dynamic people in a shopping area. There are several people present, including a person in an orange jacket walking ahead and slightly to the left, and another person in a red jacket standing near a flower planter on the right in the immediate vicinity. The pathway is clear, but it curves slightly to the right. To navigate safely, you should continue moving straight while gradually turning to the right to follow the pathway. Ensure to maintain a safe distance from the person in the orange jacket and avoid any sudden movements that could cause collisions.
**Short Text Caption ($T_s$):** People in orange and red jackets walking and standing in a busy shopping area. Curving pathway with flower planter on the right.

**Fig. 2.** Example of a long and short text caption generated by the large VLM for training *SC-CLIP*. The long text caption describes the social scenario, objects and people in the environment and the high-level navigation action for the robot. Image is from the MuSoHu dataset [29].

The trained $(\mathcal{F}_I, \mathcal{F}_T)$ are used to incorporate social and environment context to be used for trajectory planning and selection. Specifically, $\mathcal{F}_I$ is used to generate $\boldsymbol{E}_I$, from RGB images as the robot navigates an environment. $\mathcal{F}_T$ is used to create $\mathcal{D}_T$, of size $|\mathcal{D}_T|$, from an offline dataset (discussed in Section V). During navigation, $\boldsymbol{E}_I$ is used for: 1) trajectory planning by *TPN*, and 2) retrieving $\boldsymbol{E}_T$ from $\mathcal{D}_T$ based on a cosine similarity score, for trajectory selection by *TSM*.

*2) Lifelong Learning Updater (LLU)*

The objective of the *LLU* is to update $\mathcal{F}_I$ and $\mathcal{F}_T$ during robot navigation to account for new social scenarios that were not present during training. *LLU* collects a batch, $\mathcal{B}_I$, of $I_{RGB}$ during navigation and stores the batch in a buffer of size $|\mathcal{B}|$, which is a tunable parameter that determines the frequency of model updates. These images are passed to a large VLM (GPT4) to obtain the batch, $\mathcal{B}_T$, which is also used to update $\mathcal{F}_I$ and $\mathcal{F}_T$. A Symmetric Image-Text fine-tuning strategy is used to update $\mathcal{F}_I$ and $\mathcal{F}_T$ using the loss function [38]:

$$\mathcal{L}_{LLU} = - \sum_{\boldsymbol{E}_I \in V_I} \log \frac{\exp\left(\frac{\hbar(\boldsymbol{E}_I, \boldsymbol{E}_T)}{\mu}\right)}{\sum_{\boldsymbol{E}_T' \in V_T} \exp\left(\frac{\hbar(\boldsymbol{E}_I, \boldsymbol{E}_T')}{\mu}\right)}, \quad (4)$$

where $V_I$ are image features from batch $\mathcal{B}_I$, and $V_T$ are the text features from the batch $\mathcal{B}_T$, $\hbar(\cdot,\cdot)$ measures the cosine similarity score, and $\mu$ is the temperature to control the sharpness of the distribution. We denote $\mathcal{F}_I$ at update iteration $i$, as $\mathcal{F}_I^{(i)}$, and $\mathcal{F}_T$ at iteration $i$, as $\mathcal{F}_T^{(i)}$. The last iteration is denoted as $\mathcal{F}_I^{(i_t)}$ and $\mathcal{F}_T^{(i_t)}$, where $i_t$ represents the latest *LLU* iteration update. The updated encoders, $\mathcal{F}_I^{(i_t)}$ and $\mathcal{F}_T^{(i_t)}$, are used to generate $\boldsymbol{E}_I$ and $\mathcal{D}_T$, respectively.

*B. Trajectory Planning Network (TPN)*

We propose a novel *TPN* to generate socially compliant trajectory candidates, $\boldsymbol{\tau}^P$, by uniquely incorporating LiDAR data $L_{(x,y,z)}$, navigation goal $G = (x_G, y_G, \phi_G)$, and the social context embedding, $\boldsymbol{E}_I$. The *TPN* consists of a spatial context encoder backbone (*SCEB*), a multi-head attention block (*MHAB*), and a trajectory forecast head (*TFH*), Fig. 1.

The *SCEB* consists of two encoders. A LiDAR encoder to extract voxel features for the 3D points residing in each voxel [39] using five residual blocks [40] in order to obtain the LiDAR embedding vector, $\boldsymbol{E}_L$. A goal encoder uses a single layer feed-forward network (FFN) and rectified linear unit (ReLU) activation to obtain the goal embedding vector, $\boldsymbol{E}_G$.

$\boldsymbol{E}_I$, $\boldsymbol{E}_L$ and $\boldsymbol{E}_G$ are fused together using the *MHAB* to exchange context-relevant information across each embedding representation, as detailed below. The attention process starts with a randomly initialized query, $\boldsymbol{Q}^{(0)}$, which passes through three cross-attention layers: 1) *SC-CLIP* cross-attention, which attends to $\boldsymbol{E}_I$ to incorporate social and environment context from $I_{RGB}$, 2) LiDAR cross-attention, which attends to $\boldsymbol{E}_L$ to incorporate geometric and motion features from $L_{(x,y,z)}$, and 3) goal cross-attention, which attends to $\boldsymbol{E}_G$ to condition the trajectory generation on $G$. Each attention block updates the query sequentially using [41]:

$$\boldsymbol{Q}_{att}^{(z)} = \text{LN}\left(\boldsymbol{Q}^{(z)} + \mathcal{A}(\boldsymbol{Q}^{(z)},\cdot)\right), \quad (5)$$

$$\boldsymbol{Q}^{(z+1)} = \text{LN}\left(\boldsymbol{Q}_{att}^{(z)} + \text{FFN}(\boldsymbol{Q}_{att}^{(z)})\right), \quad (6)$$

where $\boldsymbol{Q}_{att}^{(z)}$ represents the intermediate query after performing cross-attention on the $z^{th}$ attention layer, $\boldsymbol{Q}^{(z)}$ represents the query for the $z^{th}$ attention layer, $\mathcal{A}$ represents one of the three aforementioned cross-attention operations, LN denotes layer normalization. The final output, $\boldsymbol{Q}^{(3)} = \boldsymbol{Q}$, is passed into the *TFH* for trajectory planning.

In the *TFH*, the output query is fed into $N_{GRU}$ GRUs with each predicting one trajectory, $\tau_k^P$. The outputs are concatenated into a single vector, $\boldsymbol{\tau}^P$, Eq. 1. The *TPN* is trained using the WTA loss function, Eq. 2, to predict multiple trajectories. The *TSM* then uses $\boldsymbol{\tau}^P$ for trajectory selection.

*C. Trajectory Selection Module (TSM)*

The objective of the *TSM* is to select a socially compliant trajectory from $\boldsymbol{\tau}^P$ using $\boldsymbol{E}_T$. Namely, $\boldsymbol{E}_T$ is passed through an FFN to produce $\boldsymbol{E}_\tau$, while $\boldsymbol{\tau}^P$ is processed by a separate FFN and GRU to produce embedding vector $\boldsymbol{E}_C$. The *TSM* is trained using the following loss function:

$$\mathcal{L}_{TSM} = \text{CE}(\text{FFN}(\boldsymbol{E}_C \oplus \boldsymbol{E}_\tau), k^*), \quad (7)$$

where $\oplus$ represents the concatenation operation and $k^*$ is an index that identifies the trajectory, $\boldsymbol{\tau}^*$, described by the navigation action in $T_l$. $\boldsymbol{\tau}^*$ is used by the *NC* for execution.

*D. Navigation Controller (NC)*

The selected $\boldsymbol{\tau}^*$, is used by *NC* to generate robot linear and angular velocities, $(v, \omega)$ using a PID controller [42].

## V. DATASETS

We collected three datasets to train the modules of OLiVia-Nav: 1) Social Context Dataset, $\mathcal{D}_{SC}$, using images from MuSoHu [29] to train *SC-CLIP*, 2) Trajectory Planning Dataset, $\mathcal{D}_{TPN}$, using expert trajectories, LiDAR point clouds and RGB images from MuSoHu to train the *TPN*, and 3) Trajectory Selection Dataset, $\mathcal{D}_{TSM}$, using expert trajectories from MuSoHu to train the *TSM*.

**1. Social Context Dataset, $\mathcal{D}_{SC}$:** This dataset comprises of 20,000 RGB images, and corresponding $(T_l, T_s)$, which are generated using GPT4o [32]. We use GPT4o for its ability to generate socially related and contextually relevant captions [43]. The text captions include descriptions of the social scenario, objects and people in the scene, and the high-level navigation plan. The database, $\mathcal{D}_T$, used within the *LLU*, is curated by taking a randomized subset of $T_l$ from $\mathcal{D}_{SC}$, and generating the corresponding $\boldsymbol{E}_T$ using $\mathcal{F}_T^{(i_t)}$.

**2. Trajectory Planning Dataset, $\mathcal{D}_{TPN}$:** This dataset comprises of 5,000 expert trajectories, LiDAR point clouds, and RGB images. The expert trajectories, $\boldsymbol{\tau}^E$, are represented by a sequence of 10 future robot poses $\{(x_i^E, y_i^E, \theta_i^E)\}_{i=0}^{10}$ from the current robot pose. For each sequence, all waypoints are transformed into the reference frame of the initial robot pose, $(x_0^E, y_0^E, \theta_0^E)$, for normalization and consistency in outputs. $I_{RGB}$, is taken at the robot's initial pose.

**3. Trajectory Selection Dataset, $\mathcal{D}_{TSM}$:** This dataset consists of 20,000 predicted trajectories, $\boldsymbol{\tau}^P$, along with the corresponding RGB images, $I_{RGB}$, $T_l$, $\boldsymbol{E}_T$, and the trajectory index, $k^*$, that corresponds to the high-level action in $T_l$.

## VI. TRAINING

OLiVia-Nav was trained in three stages: 1) *SCIE* and *SCTE* using the *SC-CLIP* framework, Eq. 3, 2) *TPN* in an end-to-end manner, Eq. 2, and 3) *TSM* based on the trajectory plans of the *TPN*, Eq. 5 and 6. Training was conducted with NVIDIA H100 GPU with 80GB of VRAM and 32GB of RAM.

### A. SC-CLIP Training

*SC-CLIP* was trained with a batch size of 256 and a cosine annealing learning rate scheduler with a learning rate (LR) of 0.0001 [44], to gradually reduce LR over time for convergence. An AdamW optimizer [45] was used with weight decay (WD) of 0.01 to prevent overfitting. *SC-CLIP* was trained for 100 epochs. We implemented early stopping and obtained the lowest validation loss epoch.

### B. TPN Training

The *TPN* was trained with a batch size of 10, LR of 0.0008, and the AdamW optimizer with WD 0.0001. $E_I$, $E_T$ and $E_G$ were projected into a common dimension of $C = 128$ using an FFN. For *MHAB*, 32 heads were used. The number of predicted trajectories was set to $K = 5$, and correspondingly $N_{GRU} = 5$. *TPN* was trained for 500 epochs until the lowest validation loss was achieved.

### C. TSM Training

The *TSM* was trained with a batch size of 128 and an LR of 0.00001. A WD of 0.00001 was used with an AdamW optimizer. *TSM* was trained for 500 epochs similar to *TPN*.

## VII. EXPERIMENTS

We conducted two sets of experiments with the Clearpath Jackal robot to evaluate the performance of OLiVia-Nav in: 1) a real-world comparison study with state-of-the-art (SOTA) social navigation methods, and 2) an ablation study to investigate the design choices of OLiVia-Nav. We used GPT4o [32] as the large VLM within the *SCM*.

### A. Comparison Study

We conducted extensive experiments in a University of Toronto building to evaluate the performance of OLiVia-Nav. Three metrics were utilized: 1) mean squared error (MSE) to determine the positional error between the robot trajectory and a ground truth trajectory, 2) Hausdorff loss [19], [46] to evaluate the shape similarity between actual and ground truth trajectories, and 3) personal space violation duration (PSV) to determine the length of time the robot travels in the personal space of a person (< 0.25 m) [47]. The ground truth trajectory was collected through teleoperation by one of our researchers using North American social navigation rules.

Four distinct social scenarios were used in the experiments, similar to [28], with 3 trials per scenario; 1) narrow hallway, where two dynamic people approach the robot front-on, Fig. 3(a), 2) blind corner, where the robot needs to approach and turn a corner while avoiding a dynamic person, Fig. 3(b), 3) navigating past two static groups and three dynamic people, Fig. 3(c), and 4) navigating through an environment with two dynamic groups and two dynamic people, Fig. 3(d).

*1) Comparison Methods:* We benchmarked OLiVia-Nav against the following SOTA methods:

- **VLM-Social-Nav** [23]**:** The **VLM-Social-Nav** method is a HMF approach which utilizes a cost-based planner with GPT4o for social navigation. The costs include: 1) obstacle collision cost obtained from LiDAR point clouds, 2) goal cost from a goal position, and 3) a social cost obtained through GPT4o from RGB images. The planner generates a set of possible robot velocity pairs, $(v, \omega)$, where the pair with the lowest combined cost is selected as the action. We choose VLM-Social-Nav to compare how the direct usage of a large VLM (GPT4o) affects navigation performance, considering its slower response time.
- **MultiSoc** [16]**:** The **MultiSoc** method is a HMB approach which predicts human trajectories by using a GNN with attention mechanisms to incorporate spatial relationships and the positions of people over time. The GNN takes as input RGB images and LiDAR point clouds to encode person positions, then uses the attention layers to predict future trajectories. The predicted trajectories are integrated into a DRL navigation policy trained with proximal policy optimization [48]. The output of the policy are robot velocity commands $(v, \omega)$. We choose MultiSoc as it is trained in a 2D simulator with procedurally generated human trajectories. This allows us to evaluate the impact of using real-world collected trajectory data for training of OLiVia-Nav.

*2) Results:* The results of the comparison study are presented in Table I. OLiVia-Nav had the lowest MSE, Hausdorff loss, and PSV across all scenarios, generating social navigation trajectories that closely match ground truth trajectories in both proximity and shape. OLiVia-Nav utilizes the *TPN*, which was trained using IL on a dataset containing real-world human trajectories and diverse environments. This enabled OLiVia-Nav to predict robot trajectories that resemble human navigation behaviors in realistic social scenarios by directly learning from real-world trajectories [9]. In contrast, VLM-Social-Nav utilized hand-tuned cost functions to generate robot trajectories for obstacle avoidance and goal reaching instead of real-world data. This approach limited the VLM-Social-Nav's ability to adapt to nuanced real-world behaviors, such as a robot dynamically adapting its speed and orientation when navigating around a corner. As a result, VLM-Social-Nav had lower performance than OLiVia-Nav. Lastly, MultiSoc resulted in robot navigations that had frequent heading direction changes, since social and environment context were not explicitly incorporated during training [16].

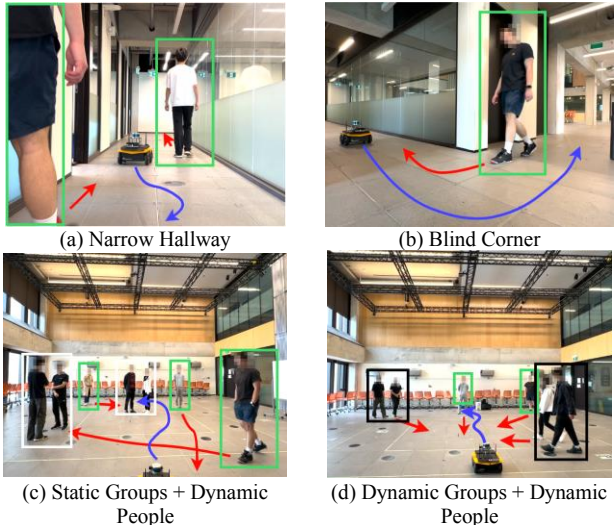

**Fig. 3.** The four experimental scenarios: (a) narrow hallway with two people, (b) approaching a blind corner, (c) static groups + dynamic people, and (d) dynamic groups + dynamic people. Red arrows represent human trajectories, blue arrows represent the robot's predicted trajectory, green boxes represent dynamic people, white boxes represent static groups, and black boxes represent dynamic groups.

TABLE I
COMPARISON RESULTS FOR THE FOUR SOCIAL SCENARIOS

| Scenario | Method | MSE↓ | Haus↓ | PSV (s)↓ |
|---|---|---|---|---|
| Narrow Hallway | OLiVia-Nav | 0.1075 | 0.7348 | 1.2 |
| | VLM-Social-Nav | 0.1915 | 0.8785 | 1.9 |
| | MultiSoc | 0.2968 | 0.9095 | 2.1 |
| Blind Corner | OLiVia-Nav | 0.0236 | 0.2572 | 0.4 |
| | VLM-Social-Nav | 0.0755 | 0.5384 | 2.6 |
| | MultiSoc | 0.1021 | 0.4596 | 2.8 |
| Static Groups + Dynamic People | OLiVia-Nav | 0.2195 | 0.7563 | 2.1 |
| | VLM-Social-Nav | 0.4361 | 1.5579 | 3.5 |
| | MultiSoc | 0.2747 | 1.1081 | 3.2 |
| Dynamic Groups + Dynamic People | OLiVia-Nav | 0.0733 | 0.4813 | 3.3 |
| | VLM-Social-Nav | 0.1459 | 0.6832 | 4.7 |
| | MultiSoc | 0.1154 | 0.7929 | 4.5 |

As the number of dynamic people in the environment increased from 1 to 6, the PSV increased for all methods due to the larger number of people encountered. In general, OLiVia-Nav demonstrated lower PSV than both VLM-Social-Nav and MultiSoc. VLM-Social-Nav had a higher PSV due to its slower response speed (average 0.6Hz), compared to OLiVia-Nav (average 5Hz), as it relied on directly querying GPT4o (large VLM) to compute social costs. This caused delays in generating navigation actions which resulted in longer durations of personal space violations. Even though MultiSoc was able to plan in real time (average 5Hz), it exhibited a higher PSV compared to OLiVia-Nav due to inaccurate human trajectory predictions. Namely, MultiSoc frequently changed the robot's heading direction in response to pose errors in the predicted human trajectories, which resulted in the robot violating personal space. Friedman tests were performed on the MSE, Hausdorff and PSV metrics for all three methods across all social scenarios. The results show statistically significant differences ($p < 0.001$). A Post-hoc Wilcoxon Signed-rank test with Bonferroni correction further confirmed statistically significant difference in all three metrics when comparing OLiVia-Nav to each method across all social scenarios ($p < 0.025$). A video of our OLiVia-Nav approach and the SOTA methods navigating a hallway (Social Scenarios 1 and 2) and open space environment (Social Scenarios 3 and 4) is on our YouTube channel: https://youtu.be/eyFJiOIITO0

*B. Ablation Study*

We conducted an ablation study to investigate the design choices of OLiVia-Nav. Namely, we considered: **1. OLiVia-Nav without (w/o) $E_I$** to determine the effect of visual semantic features from $I_{RGB}$ on trajectory planning in the *TPN*, **2. OLiVia-Nav w/o $E_T$** to evaluate the impact of token semantic features from $(T_l, T_s)$ on the selected trajectory in *TSM*, **3. OLiVia-Nav w/o $E_L$** to explore the influence of the LiDAR point cloud embedding on trajectory planning. Note that an ablation on $E_G$ does not provide useful insights since navigation requires a goal to be defined. The ablation study was conducted on a hold-out test dataset in MuSoHu.

For each of these variants, the performance was evaluated twice; Before *LLU* Update, using only $\mathcal{F}_I^{(0)}$ and $\mathcal{F}_T^{(0)}$, and After *LLU* Update, using $\mathcal{F}_I^{(i_t)}$ and $\mathcal{F}_T^{(i_t)}$, where $i_t = 1$ and $|\mathcal{B}| = 50$. The goal is to investigate the contribution of the *LLU* on all variants in terms of adapting to new social scenarios that were not present during the training of OLiVia-Nav. The performance metrics include the MSE and Hausdorff loss.

*2) Results:* The ablation study results are presented in Table II. The Before *LLU* Update results show that the OLiVia-Nav had the lowest MSE and Hausdorff loss of 0.2981 and 1.8087 compared to all variants. The After *LLU* Update results also showed OLiVia-Nav having the lowest MSE and Hausdorff loss of 0.2521 and 1.4893, respectively. All ablation variants also improved their performance with the updates. For example, OLiVia-Nav w/o $E_I$ achieved a higher MSE and Hausdorff loss compared to OLiVia-Nav as the predicted robot trajectories did not consider social and environment context. OLiVia-Nav w/o $E_T$ also achieved degraded navigation performance as the social context embedding from the token semantic features of the text captions was not used. This embedding encodes the high-level navigation action and is necessary for trajectory selection. The variant was unable to utilize high-level action information without this embedding, resulting in the random selection of trajectories. Lastly, as OLiVia-Nav w/o $E_L$ did not the use LiDAR embedding for obstacle detection, trajectory plans resulted in collisions with objects and people.

TABLE II
ABLATION STUDY

| Methods | Before *LLU* Update | | After *LLU* Update | |
|---|---|---|---|---|
| | MSE↓ | Haus↓ | MSE↓ | Haus↓ |
| OLiVia-Nav (ours) | 0.2981 | 1.8087 | 0.2521 | 1.4893 |
| OLiVia-Nav w/o $E_I$ | 0.4412 | 2.3772 | 0.3701 | 2.0839 |
| OLiVia-Nav w/o $E_T$ | 0.3838 | 2.1063 | 0.3791 | 2.0349 |
| OLiVia-Nav w/o $E_L$ | 0.4382 | 2.3681 | 0.4180 | 2.1894 |

VIII. CONCLUSION

In this paper, we present a novel lifelong vision language architecture OLiVia-Nav which uses VLMs to address the robot social navigation problem in dynamic human environments. Our approach introduces a novel distillation process, *SC-CLIP*, to leverage the social reasoning capabilities of large VLMs for trajectory planning while being able to adapt online to new scenarios using the lifelong learning ability of the lightweight VLM. Extensive real-world experiments demonstrate that OLiVia-Nav follows expert trajectories more accurately compared to SOTA methods. Furthermore, ablation studies show the importance of each of the embeddings on robot social navigation performance. Future work will investigate the performance of the OLiVia-Nav in larger environments with crowds.

## REFERENCES

[1] A. Francis *et al.*, "Principles and Guidelines for Evaluating Social Robot Navigation Algorithms," Sep. 19, 2023, *arXiv*: arXiv:2306.16740. Accessed: Jun. 18, 2024. [Online]. Available: http://arxiv.org/abs/2306.16740

[2] A. A. Morgan, J. Abdi, M. A. Q. Syed, G. E. Kohen, P. Barlow, and M. P. Vizcaychipi, "Robots in Healthcare: a Scoping Review," *Curr Robot Rep*, vol. 3, no. 4, pp. 271–280, Oct. 2022, doi: 10.1007/s43154-022-00095-4.

[3] R. Memmesheimer *et al.*, "Cleaning Robots in Public Spaces: A Survey and Proposal for Benchmarking Based on Stakeholders Interviews," Jul. 23, 2024, *arXiv*: arXiv:2407.16393. Accessed: Sep. 15, 2024. [Online]. Available: http://arxiv.org/abs/2407.16393

[4] C. Getson and G. Nejat, "Socially Assistive Robots Helping Older Adults through the Pandemic and Life after COVID-19," *Robotics*, vol. 10, no. 3, p. 106, Sep. 2021, doi: 10.3390/robotics10030106.

[5] C. Getson and G. Nejat, "The Robot Screener Will See You Now: A Socially Assistive Robot for COVID-19 Screening in Long-Term Care Homes," in *2022 31st IEEE International Conference on Robot and Human Interactive Communication (RO-MAN)*, Napoli, Italy: IEEE, Aug. 2022, pp. 672–677. doi: 10.1109/RO-MAN53752.2022.9900620.

[6] J. Wang, W. P. Chan, P. Carreno-Medrano, A. Cosgun, and E. Croft, "Metrics for Evaluating Social Conformity of Crowd Navigation Algorithms," in *2022 IEEE International Conference on Advanced Robotics and Its Social Impacts (ARSO)*, May 2022, pp. 1–6. doi: 10.1109/ARSO54254.2022.9802981.

[7] P. T. Singamaneni *et al.*, "A survey on socially aware robot navigation: Taxonomy and future challenges," *The International Journal of Robotics Research*, p. 02783649241230562, Feb. 2024, doi: 10.1177/02783649241230562.

[8] C. Getson and G. Nejat, "The adoption of socially assistive robots for long-term care: During COVID-19 and in a post-pandemic society," *Healthc Manage Forum*, vol. 35, no. 5, pp. 301–309, Sep. 2022, doi: 10.1177/08404704221106406.

[9] C. Mavrogiannis *et al.*, "Core Challenges of Social Robot Navigation: A Survey," *J. Hum.-Robot Interact.*, vol. 12, no. 3, pp. 1–39, Sep. 2023, doi: 10.1145/3583741.

[10] R. Mirsky, X. Xiao, J. Hart, and P. Stone, "Conflict Avoidance in Social Navigation -- a Survey," Dec. 28, 2022, *arXiv*: arXiv:2106.12113. Accessed: Apr. 05, 2024. [Online]. Available: http://arxiv.org/abs/2106.12113

[11] W. Wang, R. Wang, L. Mao, and B.-C. Min, "NaviSTAR: Socially Aware Robot Navigation with Hybrid Spatio-Temporal Graph Transformer and Preference Learning," Sep. 26, 2023, *arXiv*: arXiv:2304.05979. Accessed: Mar. 31, 2024. [Online]. Available: http://arxiv.org/abs/2304.05979

[12] A. J. Sathyamoorthy, U. Patel, M. Paul, N. K. S. Kumar, Y. Savle, and D. Manocha, "CoMet: Modeling Group Cohesion for Socially Compliant Robot Navigation in Crowded Scenes," *IEEE Robot. Autom. Lett.*, vol. 7, no. 2, pp. 1008–1015, Apr. 2022, doi: 10.1109/LRA.2021.3135560.

[13] V. Narayanan, B. M. Manoghar, R. P. RV, and A. Bera, "EWareNet: Emotion Aware Human Intent Prediction and Adaptive Spatial Profile Fusion for Social Robot Navigation," Mar. 07, 2023, *arXiv*: arXiv:2011.09438. Accessed: Apr. 08, 2024. [Online]. Available: http://arxiv.org/abs/2011.09438

[14] L. Kästner, J. Li, Z. Shen, and J. Lambrecht, "Enhancing Navigational Safety in Crowded Environments using Semantic-Deep-Reinforcement-Learning-based Navigation," 2021, doi: 10.48550/ARXIV.2109.11288.

[15] N. Hirose, D. Shah, A. Sridhar, and S. Levine, "SACSoN: Scalable Autonomous Control for Social Navigation," *IEEE Robot. Autom. Lett.*, vol. 9, no. 1, pp. 49–56, Jan. 2024, doi: 10.1109/LRA.2023.3329626.

[16] E. Escudie, L. Matignon, and J. Saraydaryan, "Attention Graph for Multi-Robot Social Navigation with Deep Reinforcement Learning," Jan. 31, 2024, *arXiv*: arXiv:2401.17914. Accessed: Jul. 29, 2024. [Online]. Available: http://arxiv.org/abs/2401.17914

[17] J. Liang, U. Patel, A. J. Sathyamoorthy, and D. Manocha, "Crowd-Steer: Realtime Smooth and Collision-Free Robot Navigation in Densely Crowded Scenarios Trained using High-Fidelity Simulation," in *Proceedings of the Twenty-Ninth International Joint Conference on Artificial Intelligence*, Yokohama, Japan: International Joint Conferences on Artificial Intelligence Organization, Jul. 2020, pp. 4221–4228. doi: 10.24963/ijcai.2020/583.

[18] J. Xu, W. Zhang, J. Cai, and H. Liu, "SafeCrowdNav: safety evaluation of robot crowd navigation in complex scenes," *Front. Neurorobot.*, vol. 17, p. 1276519, Oct. 2023, doi: 10.3389/fnbot.2023.1276519.

[19] A. H. Raj *et al.*, "Rethinking Social Robot Navigation: Leveraging the Best of Two Worlds," Mar. 09, 2024, *arXiv*: arXiv:2309.13466. Accessed: Apr. 05, 2024. [Online]. Available: http://arxiv.org/abs/2309.13466

[20] B. Panigrahi, A. H. Raj, M. Nazeri, and X. Xiao, "A Study on Learning Social Robot Navigation with Multimodal Perception," Sep. 21, 2023, *arXiv*: arXiv:2309.12568. Accessed: Apr. 05, 2024. [Online]. Available: http://arxiv.org/abs/2309.12568

[21] L. Tai, J. Zhang, M. Liu, and W. Burgard, "Socially Compliant Navigation through Raw Depth Inputs with Generative Adversarial Imitation Learning," Feb. 26, 2018, *arXiv*: arXiv:1710.02543. doi: 10.48550/arXiv.1710.02543.

[22] H. Kretzschmar, M. Spies, C. Sprunk, and W. Burgard, "Socially compliant mobile robot navigation via inverse reinforcement learning," *The International Journal of Robotics Research*, vol. 35, no. 11, pp. 1289–1307, Sep. 2016, doi: 10.1177/0278364915619772.

[23] D. Song, J. Liang, A. Payandeh, X. Xiao, and D. Manocha, "Socially Aware Robot Navigation through Scoring Using Vision-Language Models," Mar. 29, 2024, *arXiv*: arXiv:2404.00210. Accessed: Apr. 05, 2024. [Online]. Available: http://arxiv.org/abs/2404.00210

[24] W. Wang, L. Mao, R. Wang, and B.-C. Min, "SRLM: Human-in-Loop Interactive Social Robot Navigation with Large Language Model and Deep Reinforcement Learning," Mar. 22, 2024, *arXiv*: arXiv:2403.15648. Accessed: Mar. 31, 2024. [Online]. Available: http://arxiv.org/abs/2403.15648

[25] C. Weinrich, M. Volkhardt, E. Einhorn, and H.-M. Gross, "Prediction of human collision avoidance behavior by lifelong learning for socially compliant robot navigation," in *2013 IEEE International Conference on Robotics and Automation*, Karlsruhe, Germany: IEEE, May 2013, pp. 376–381. doi: 10.1109/ICRA.2013.6630603.

[26] I. Okunevich, A. Lombard, T. Krajnik, Y. Ruichek, and Z. Yan, "Online Context Learning for Socially-compliant Navigation," Jun. 17, 2024, *arXiv*: arXiv:2406.11495. Accessed: Jul. 16, 2024. [Online]. Available: http://arxiv.org/abs/2406.11495

[27] J. Redmon and A. Farhadi, "YOLOv3: An Incremental Improvement," Apr. 08, 2018, *arXiv*: arXiv:1804.02767. Accessed: Sep. 14, 2024. [Online]. Available: http://arxiv.org/abs/1804.02767

[28] H. Karnan *et al.*, "Socially Compliant Navigation Dataset (SCAND): A Large-Scale Dataset of Demonstrations for Social Navigation," Jun. 08, 2022, *arXiv*: arXiv:2203.15041. Accessed: Apr. 05, 2024. [Online]. Available: http://arxiv.org/abs/2203.15041

[29] D. M. Nguyen, M. Nazeri, A. Payandeh, A. Datar, and X. Xiao, "Toward Human-Like Social Robot Navigation: A Large-Scale, Multi-Modal, Social Human Navigation Dataset," Aug. 09, 2023, *arXiv*: arXiv:2303.14880. Accessed: Apr. 05, 2024. [Online]. Available: http://arxiv.org/abs/2303.14880

[30] T. Schreiter *et al.*, "TH\"OR-MAGNI: A Large-scale Indoor Motion Capture Recording of Human Movement and Robot Interaction," Mar. 14, 2024, *arXiv*: arXiv:2403.09285. Accessed: Sep. 13, 2024. [Online]. Available: http://arxiv.org/abs/2403.09285

[31] W. Chen, O. Mees, A. Kumar, and S. Levine, "Vision-Language Models Provide Promptable Representations for Reinforcement Learning," Feb. 13, 2024, *arXiv*: arXiv:2402.02651. Accessed: Apr. 12, 2024. [Online]. Available: http://arxiv.org/abs/2402.02651

[32] OpenAI *et al.*, "GPT-4 Technical Report," Mar. 04, 2024, *arXiv*: arXiv:2303.08774. Accessed: Jul. 29, 2024. [Online]. Available: http://arxiv.org/abs/2303.08774

[33] S. Kim, H. Jeon, J. Choi, and D. Kum, "Diverse Multiple Trajectory Prediction Using a Two-stage Prediction Network Trained with Lane Loss," *IEEE Robot. Autom. Lett.*, vol. 8, no. 4, pp. 2038–2045, Apr. 2023, doi: 10.1109/LRA.2022.3231525.

[34] B. Zhang, P. Zhang, X. Dong, Y. Zang, and J. Wang, "Long-CLIP: Unlocking the Long-Text Capability of CLIP," May 23, 2024, *arXiv*: arXiv:2403.15378. Accessed: May 29, 2024. [Online]. Available: http://arxiv.org/abs/2403.15378

[35] A. Radford *et al.*, "Learning Transferable Visual Models From Natural Language Supervision," Feb. 26, 2021, *arXiv*:

arXiv:2103.00020. Accessed: Jun. 23, 2024. [Online]. Available: http://arxiv.org/abs/2103.00020

[36] A. Vaswani *et al.*, "Attention Is All You Need," Aug. 01, 2023, *arXiv*: arXiv:1706.03762. Accessed: Jun. 26, 2024. [Online]. Available: http://arxiv.org/abs/1706.03762

[37] J. Shlens, "A Tutorial on Principal Component Analysis," Apr. 03, 2014, *arXiv*: arXiv:1404.1100. Accessed: Jul. 29, 2024. [Online]. Available: http://arxiv.org/abs/1404.1100

[38] L. Wang, L. Xiang, Y. Wei, Y. Wang, and Z. He, "CLIP model is an Efficient Online Lifelong Learner," May 23, 2024, *arXiv*: arXiv:2405.15155. Accessed: Jun. 19, 2024. [Online]. Available: http://arxiv.org/abs/2405.15155

[39] Y. Zhou and O. Tuzel, "VoxelNet: End-to-End Learning for Point Cloud Based 3D Object Detection," in *2018 IEEE/CVF Conference on Computer Vision and Pattern Recognition*, Salt Lake City, UT, USA: IEEE, Jun. 2018, pp. 4490–4499. doi: 10.1109/CVPR.2018.00472.

[40] K. He, X. Zhang, S. Ren, and J. Sun, "Deep Residual Learning for Image Recognition," Dec. 10, 2015, *arXiv*: arXiv:1512.03385. Accessed: Jul. 29, 2024. [Online]. Available: http://arxiv.org/abs/1512.03385

[41] S. Casas *et al.*, "DeTra: A Unified Model for Object Detection and Trajectory Forecasting," Jun. 13, 2024, *arXiv*: arXiv:2406.04426. Accessed: Jun. 15, 2024. [Online]. Available: http://arxiv.org/abs/2406.04426

[42] Kiam Heong Ang, G. Chong, and Yun Li, "PID control system analysis, design, and technology," *IEEE Trans. Contr. Syst. Technol.*, vol. 13, no. 4, pp. 559–576, Jul. 2005, doi: 10.1109/TCST.2005.847331.

[43] K. Gandhi, J.-P. Fränken, T. Gerstenberg, and N. D. Goodman, "Understanding Social Reasoning in Language Models with Language Models," Dec. 04, 2023, *arXiv*: arXiv:2306.15448. Accessed: Sep. 15, 2024. [Online]. Available: http://arxiv.org/abs/2306.15448

[44] I. Loshchilov and F. Hutter, "SGDR: Stochastic Gradient Descent with Warm Restarts," May 03, 2017, *arXiv*: arXiv:1608.03983. Accessed: Aug. 23, 2024. [Online]. Available: http://arxiv.org/abs/1608.03983

[45] I. Loshchilov and F. Hutter, "Decoupled Weight Decay Regularization," Jan. 04, 2019, *arXiv*: arXiv:1711.05101. Accessed: Aug. 23, 2024. [Online]. Available: http://arxiv.org/abs/1711.05101

[46] D. Kraft, "Computing the Hausdorff Distance of Two Sets from Their Signed Distance Functions," *Int. J. Comput. Geom. Appl.*, vol. 30, no. 01, pp. 19–49, Mar. 2020, doi: 10.1142/S0218195920500028.

[47] N. Hirose, D. Shah, K. Stachowicz, A. Sridhar, and S. Levine, "SELFI: Autonomous Self-Improvement with Reinforcement Learning for Social Navigation," Mar. 01, 2024, *arXiv*: arXiv:2403.00991. Accessed: Mar. 13, 2024. [Online]. Available: http://arxiv.org/abs/2403.00991

[48] C. Yu *et al.*, "The Surprising Effectiveness of PPO in Cooperative, Multi-Agent Games," Nov. 04, 2022, *arXiv*: arXiv:2103.01955. Accessed: Sep. 09, 2024. [Online]. Available: http://arxiv.org/abs/2103.01955